\documentclass[sigconf,nonacm]{acmart}
\setcopyright{none}
\renewcommand\footnotetextcopyrightpermission[1]{}

\usepackage{xurl}
\usepackage{graphicx}
\usepackage{algorithm}
\usepackage{algorithmic}
\usepackage{amsmath}
\usepackage{amsthm}
\usepackage{booktabs}
\usepackage{multirow}
\usepackage{newfloat}
\usepackage{listings}
\usepackage{tabularx}

\DeclareCaptionStyle{ruled}{labelfont=normalfont,labelsep=colon,strut=off}
\floatstyle{ruled}
\newfloat{listing}{tb}{lst}{}
\floatname{listing}{Listing}

\title{STKAN: Kolmogorov-Arnold Networks for Spatio-Temporal Forecasting}

\author{Sicong Lai}
\affiliation{%
  \institution{The Hong Kong University of Science and Technology (Guangzhou)}
  \country{China}}
\author{Yuehong Hu}
\affiliation{%
  \institution{The Hong Kong University of Science and Technology (Guangzhou)}
  \country{China}}
\author{Siru Zhong}
\affiliation{%
  \institution{The Hong Kong University of Science and Technology (Guangzhou)}
  \country{China}}
\author{Si Qiao}
\affiliation{%
  \institution{The Hong Kong University of Science and Technology (Guangzhou)}
  \country{China}}
\author{Yuxuan Liang}
\affiliation{%
  \institution{The Hong Kong University of Science and Technology (Guangzhou)}
  \country{China}}
\author{Guangyin Jin}
\authornote{Corresponding author.}
\affiliation{%
  \institution{Chang'an University}
  \country{China}}
\affiliation{%
  \institution{China University of Geosciences}
  \country{China}}

\begin{document}

\begin{abstract}
Real-world traffic data exhibit heterogeneous spatial correlations and nonlinear temporal dynamics, posing substantial challenges for accurate spatio-temporal forecasting. Existing approaches have developed increasingly sophisticated graph, attention, and decomposition architectures, while the influence of the underlying nonlinear function approximator has received comparatively less attention. In this work, we propose STKAN, a spatio-temporal forecasting architecture that introduces Taylor-polynomial Kolmogorov--Arnold Network modules into spatial and temporal token mixing. STKAN first constructs high-level spatial representations through a learnable soft node-group assignment mechanism, applies group-wise spatial mixing, and subsequently models temporal dependencies over the compressed sequence. Spatial and temporal self-attention layers are further employed to capture long-range interactions. Experiments on five traffic forecasting benchmarks show that STKAN achieves competitive performance and performs better than the evaluated MLP-based variant in the tested settings. These results suggest that the design of nonlinear function approximators can serve as a useful complement to architectural design in spatio-temporal forecasting.
\end{abstract}

\maketitle

\section{Introduction}

Spatio-temporal forecasting serves as a foundational pillar within the AI4Transport ecosystem. It drives intelligent transportation technologies \cite{1,2,3}. By predicting future road traffic conditions from historical data, AI agents enable services such as autonomous traffic management, route optimization, and road safety support \cite{4,5,6}. Traffic signals exhibit spatial heterogeneity and complex temporal dynamics. Spatially, interactions between regions are intricate and non-Euclidean. Temporally, signals exhibit strong periodicity entangled with abrupt changes and non-stationary patterns. These properties motivate forecasting models that can capture both spatial dependencies and temporal variations.

\begin{figure}[t]
  \centering
  \includegraphics[width=0.95\linewidth]{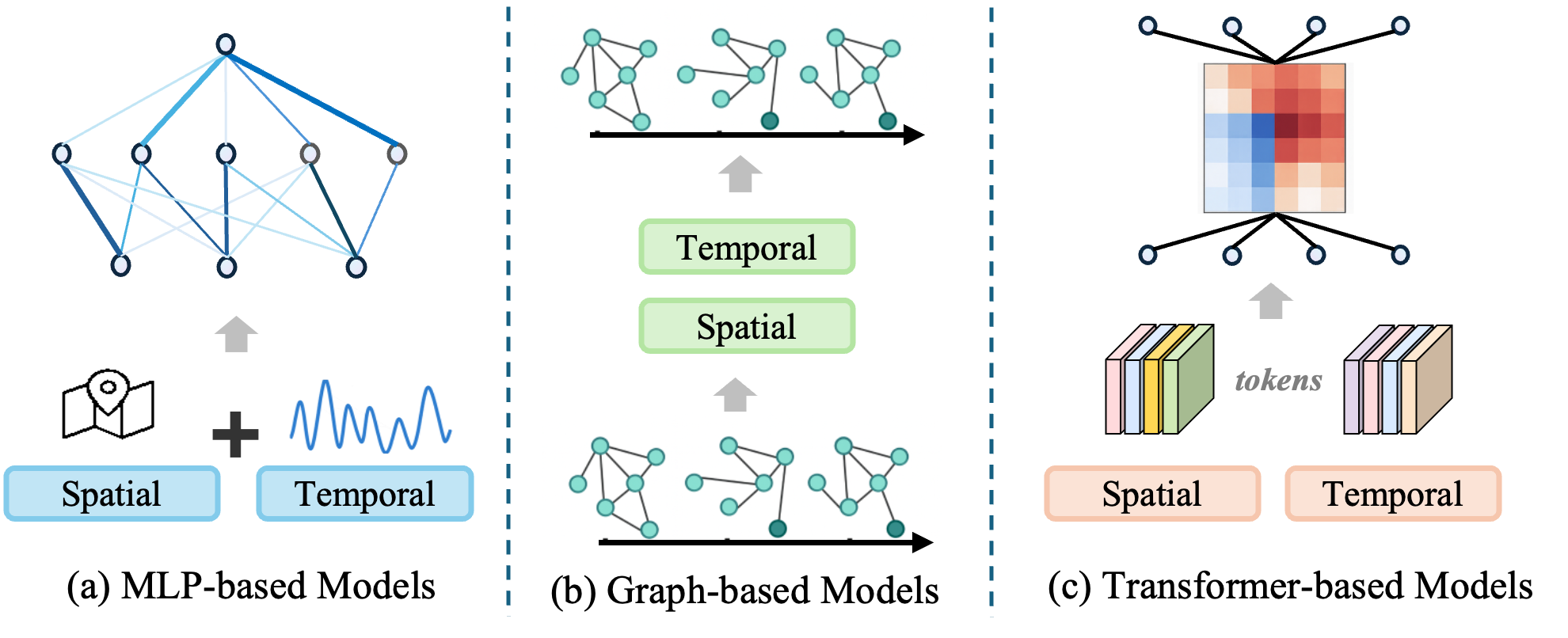}
  \caption{Illustration of different spatio-temporal models.}
  \label{fig:11}
\end{figure}

To address these challenges, researchers have developed sophisticated architectures to explicitly model these dependencies. As illustrated in Figure~\ref{fig:11}, Graph-based methods perform forecasting by assuming predefined or learned graph structures. These structures capture spatial correlations among nodes \cite{jin2023spatio}. Representative works such as STGCN \cite{yu2017spatio} and Graph WaveNet \cite{wu2019graph} leverage graph convolutions and adaptive adjacency matrices. Meanwhile, Transformer-based methods utilize attention mechanisms. They model long-range dependencies across spatio-temporal tokens \cite{staeformer}. Additionally, spectral decomposition techniques like StemGNN \cite{stemgnn} and STWave \cite{stwave} focus on frequency-domain transforms. Frameworks like STID \cite{shao2022spatial} utilize identity-based embeddings. These decoupled methods aim to obtain latent spatio-temporal representations that support more accurate predictions.

Despite diverse structural designs, many spatio-temporal architectures employ MLPs with fixed activation functions as their default nonlinear mapping modules. Although MLPs are expressive general-purpose approximators, their nonlinearities are usually selected in advance and shared across units. This observation motivates a complementary research question: beyond architectural design, to what extent does the choice of nonlinear function approximator affect spatio-temporal forecasting performance?

In this work, we investigate whether the choice of nonlinear function approximator constitutes an underexplored factor in spatio-temporal forecasting performance. We propose STKAN, a Kolmogorov--Arnold Network (KAN)-based architecture that decomposes spatial and temporal dependencies while introducing TaylorKAN layers into the spatial and temporal token-mixing mappings. STKAN provides a controlled architectural setting for examining whether Taylor-polynomial KAN token mixers can complement conventional spatio-temporal modeling components. The channel mixer and Transformer feed-forward network remain MLP-based, so the model does not replace all MLP components.

Our main contributions are summarized as follows:

\begin{itemize}
\item We investigate the role of nonlinear function-approximator design in spatio-temporal forecasting and introduce STKAN, which incorporates Taylor-polynomial KAN mappings into spatial and temporal token-mixing modules.
\item We develop a learnable soft node-group assignment mechanism together with dedicated spatial and temporal mixing blocks. The grouping mechanism constructs compact spatial representations, while the two token mixers model interactions along the spatial-group and temporal dimensions, respectively.
\item We evaluate STKAN on five traffic forecasting benchmarks. The results demonstrate competitive forecasting accuracy, while the existing ablation results suggest that the TaylorKAN token mixers, adaptive spatial grouping, and attention components each contribute to the overall model in the evaluated configurations.
\end{itemize}

\section{Related Work}

\subsection{Spatio-Temporal Forecasting}

Spatio-temporal forecasting extends traditional time-series forecasting by incorporating both temporal dynamics and spatial dependencies, such as in traffic management, where multiple traffic sensors' data is used to predict future conditions. Early deep learning approaches combined Convolutional Neural Networks (CNNs) and Recurrent Neural Networks (RNNs) to capture spatial and temporal dependencies \cite{shi2015convolutional,yao2018deep,lai2018modeling}. However, grid-based CNNs may not effectively handle non-Euclidean spatial relationships, leading to the development of Graph Convolutional Networks (GCNs) \cite{defferrard2016convolutional,kipf2016semi} and spatio-temporal Graph Neural Networks (STGNNs) \cite{li2017diffusion,yu2017spatio,li2023dynamic}. These models, such as DCRNN \cite{li2017diffusion}, ST-MetaNet \cite{pan2019urban}, and DGCRN \cite{li2023dynamic}, integrate GCNs with RNNs \cite{cho2014properties}, while others like Graph WaveNet \cite{wu2019graph} and STGCN \cite{yu2017spatio} combine GCNs with gated Temporal Convolutional Networks (TCNs). Attention mechanisms have also been widely adopted in STGNNs \cite{zheng2020gman}. Some studies criticize the reliance on pre-defined graphs, suggesting alternatives like AGCRN \cite{bai2020adaptive} and MTGNN \cite{wu2020connecting}, which learn latent graph structures. Recent non-GCN solutions, such as STNorm \cite{deng2021st} and STID \cite{shao2022spatial}, further highlight the need for a better understanding of spatial dependencies in forecasting tasks.

\subsection{Kolmogorov-Arnold Network}

KANs are motivated by the Kolmogorov--Arnold representation theorem \cite{liu2024kan}, which represents multivariate functions through compositions of univariate functions. Building upon this paradigm, several variants have been proposed, including wavelet-based WavKAN \cite{Wav-kan}, Taylor polynomial-based TaylorKAN \cite{taylorkan}, fKAN with trainable Jacobi basis functions \cite{jacobi}, and FastKAN \cite{fastkan}, which employs radial basis function approximations of B-spline bases.

Although KANs are motivated by the Kolmogorov--Arnold representation theorem, the theorem itself does not imply that a finite KAN model will universally outperform an MLP in empirical forecasting tasks. In this work, KANs are therefore treated as an alternative nonlinear parameterization with a different inductive bias, rather than as a theoretically guaranteed replacement for MLPs.

KANs have recently attracted attention in time series forecasting. Prior work has explored KAN-based models for time series analysis \cite{kants}, mixture-of-experts KAN designs \cite{han2024kans}, symbolic-regression-oriented T-KAN and MT-KAN models \cite{tkanmtkan}, collaborative time-frequency learning with iTFKAN \cite{iTFKAN}, and frequency-decomposition learning with TimeKAN \cite{huang2025timekan}. BiLSTM-KAN integrates KAN layers with a bidirectional LSTM backbone for traffic flow forecasting \cite{BiLSTM}. STKAN differs from these studies by replacing the token-mixing mappings in its spatial and temporal mixer blocks with Taylor-polynomial KAN layers, while retaining conventional MLPs in the channel-mixing and Transformer feed-forward components.

\section{Preliminary}

Spatio-temporal forecasting is a specialized multivariate time-series forecasting problem. Given a historical input sequence $\mathbf{X}_{t-L_{\mathrm{in}}+1:t}=[\mathbf{X}_{t-L_{\mathrm{in}}+1},\ldots,\mathbf{X}_{t}]\in\mathbb{R}^{L_{\mathrm{in}}\times N\times C_{\mathrm{in}}}$, the goal is to predict $\widehat{\mathbf{Y}}_{t+1:t+H}=[\widehat{\mathbf{Y}}_{t+1},\ldots,\widehat{\mathbf{Y}}_{t+H}]\in\mathbb{R}^{H\times N\times C_{\mathrm{out}}}$.
Here, $L_{\mathrm{in}}$ denotes the historical input length, $H$ denotes the prediction horizon, $N$ is the number of spatial nodes, and $C_{\mathrm{in}}$ and $C_{\mathrm{out}}$ are the input and output channel dimensions.

\section{Methodology}

\begin{figure*}[!t]  
    \centering
    \includegraphics[width=0.8\textwidth]{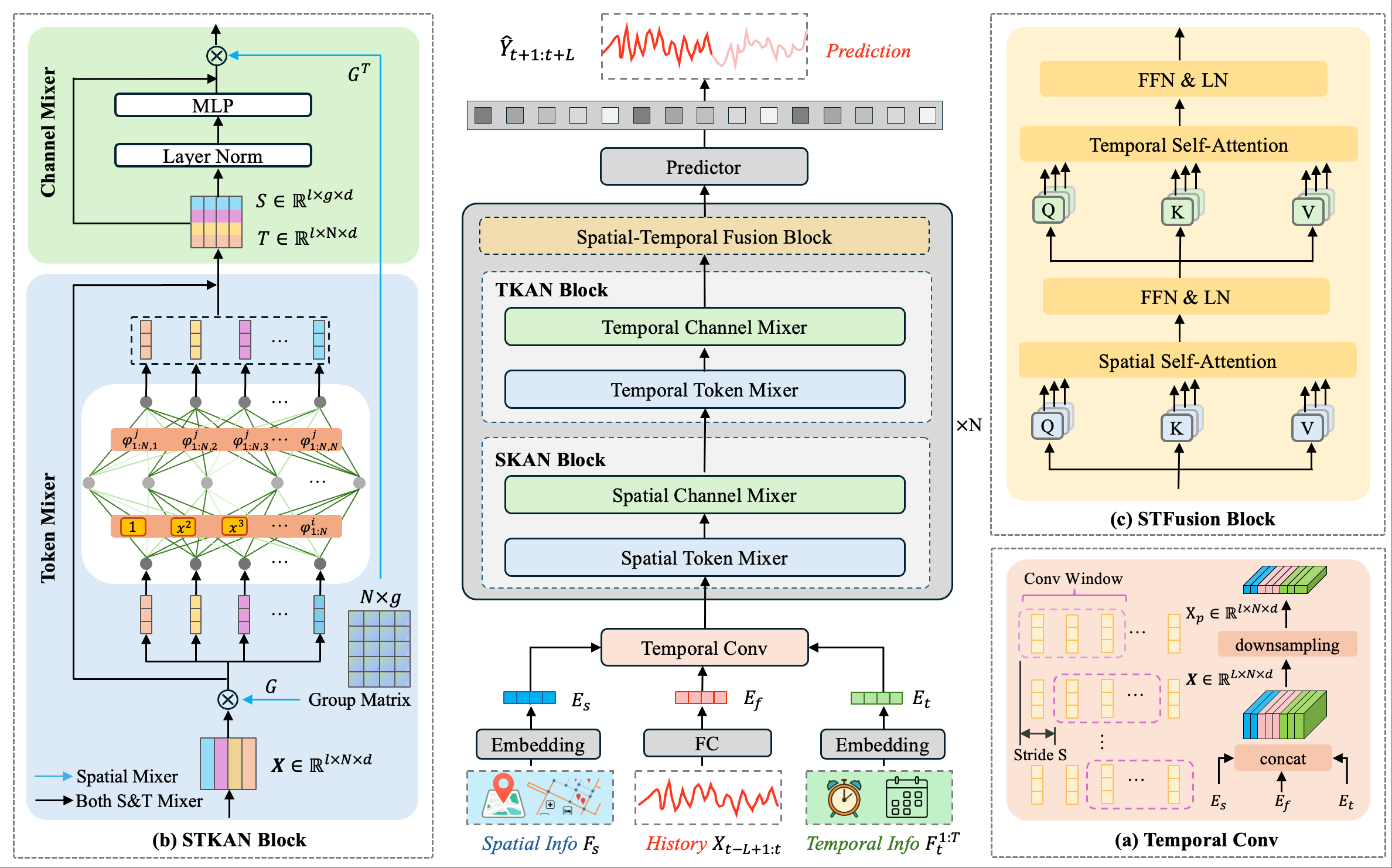} 
    \caption{Overview of our proposed STKAN framework.}
    \label{fig:1}
\end{figure*}

In this paper, we propose STKAN to model spatial interactions and temporal dynamics by introducing TaylorKAN token mixers into a spatio-temporal forecasting architecture. The overall architecture of STKAN is shown in Figure~\ref{fig:1}.

\subsection{Embedding Layers}

To capture spatio-temporal dependencies in traffic sequences, we embed the raw input $\mathbf{X}\in\mathbb{R}^{L_{\mathrm{in}}\times N\times C_{\mathrm{in}}}$ with a fully connected layer as $\mathbf{E}_f=\mathrm{FC}(\mathbf{X})\in\mathbb{R}^{L_{\mathrm{in}}\times N\times d_f}$. We use a learnable spatial embedding $\mathbf{E}_s\in\mathbb{R}^{N\times d_s}$.
The spatial embedding is shared across time steps and is broadcast along the temporal dimension before concatenation. To incorporate temporal periodicity, we use two embedding dictionaries: a time-of-day embedding with $T_{\mathrm{tod}}=288$ slots for five-minute data and a day-of-week embedding with $T_{\mathrm{dow}}=7$ categories. The resulting embeddings are denoted as $\mathbf{E}_{\mathrm{tod}}$ and $\mathbf{E}_{\mathrm{dow}}$ after lookup and broadcasting to the node dimension.

\subsection{Temporal Convolution Blocks}

All embeddings are concatenated along the feature dimension to form the initial representation:
\begin{equation}
\mathbf{X}^{(0)}
=
\mathbf{E}_f
\Vert
\mathbf{E}_s
\Vert
\mathbf{E}_{\mathrm{tod}}
\Vert
\mathbf{E}_{\mathrm{dow}}
\in
\mathbb{R}^{L_{\mathrm{in}}\times N\times d_h},
\label{eq:initial_embedding}
\end{equation}
where $d_h=d_f+d_s+d_{\mathrm{tod}}+d_{\mathrm{dow}}$. To capture local temporal context while reducing the temporal resolution, we define a patch-extraction operator that applies a $1\times w$ convolution along the time axis with stride $s$:
\begin{equation}
\mathbf{X}_p
=
\mathrm{PatchConv}
\left(
\mathbf{X}^{(0)};w,s
\right)
\in
\mathbb{R}^{L_p\times N\times d_h},
\label{eq:patch_conv}
\end{equation}
where $L_p=\left\lfloor\frac{L_{\mathrm{in}}-w}{s}\right\rfloor+1$.
Thus, $\mathbf{X}_p$ serves as a patch-level spatio-temporal feature map for subsequent mixer and attention blocks.

\subsection{Spatio-Temporal KAN Blocks}
The STKAN blocks combine learnable spatial grouping, spatial TaylorKAN token mixing, and temporal TaylorKAN token mixing. STKAN first uses a soft assignment matrix to aggregate raw nodes into macro-level spatial tokens. The aggregated representations are then processed by the Spatial KAN (SKAN) block. The spatial output is mapped back to node-level features and passed to the Temporal KAN (TKAN) block, which models dependencies over the compressed temporal dimension.

\textbf{Learnable Spatial Grouping.} To transform the original spatial nodes into macro-level spatial tokens, we introduce an unnormalized learnable parameter matrix $\mathbf{A}\in\mathbb{R}^{N\times G}$.
The assignment matrix is obtained by applying softmax along the group dimension:
\begin{equation}
\mathbf{G}_{n,g}
=
\frac{\exp(\mathbf{A}_{n,g})}
{\sum_{g'=1}^{G}\exp(\mathbf{A}_{n,g'})}.
\label{eq:group_softmax}
\end{equation}
Thus, $\sum_{g=1}^{G}\mathbf{G}_{n,g}=1$ for each node $n$. Given $\mathbf{X}_p\in\mathbb{R}^{L_p\times N\times d_h}$, the group-level representation $\widetilde{\mathbf{X}}\in\mathbb{R}^{L_p\times G\times d_h}$ is computed as
\begin{equation}
\widetilde{\mathbf{X}}_{\tau,g,c}
=
\sum_{n=1}^{N}
\mathbf{G}_{n,g}
\mathbf{X}_{p,\tau,n,c}.
\label{eq:spatial_grouping}
\end{equation}

\textbf{Spatial KAN Block.}  
The SKAN block follows a mixer-style design. Although MLPs are expressive general-purpose approximators, their activation functions are usually fixed before training. KAN-based mappings provide a different parameterization in which learnable univariate functions are associated with network connections. The transmission from the $j$-th neuron in layer $\ell+1$ to all neurons in layer $\ell$ is formulated as:
\begin{equation}
    z_{\ell+1, j} = \sum_{i=1}^{n_\ell} \phi_{\ell, j, i}(z_{\ell, i}),
\end{equation}
where $z_{\ell, i}$ is the $i$-th neuron in layer $\ell$, $n_\ell$ is the number of neurons in that layer, and $\phi_{\ell, j, i}(\cdot)$ is a learnable univariate mapping.

In our model, $\phi$ is instantiated by a TaylorKAN layer. A Taylor expansion motivates the polynomial basis:
\begin{equation}
f(x)
\approx
\sum_{r=0}^{K}
\frac{f^{(r)}(0)}{r!}x^r.
\end{equation}
The implemented layer uses learnable coefficients:
\begin{equation}
\phi_q(\mathbf{x})
=
\sum_{p=1}^{C}
\sum_{r=0}^{K}
\theta_{q,p,r}x_p^r+b_q.
\label{eq:taylor_layer}
\end{equation}
The coefficients $\theta_{q,p,r}$ are optimized directly and are not constrained to equal the analytical derivatives of an underlying function. Accordingly, the adopted layer is more precisely interpreted as a Taylor-inspired polynomial-basis KAN layer. The polynomial basis allows the mapping to combine nonlinear terms of different orders. Its effectiveness for spatial and temporal mixing is evaluated empirically in the subsequent experiments.

To avoid ambiguity when mixing tensor dimensions, we define a spatial permutation operator $\mathcal{P}_s:\mathbb{R}^{L_p\times G\times d_h}\rightarrow\mathbb{R}^{L_p\times d_h\times G}$.
The spatial token-mixing stage is then
\begin{equation}
\mathbf{U}_s
=
\mathcal{P}_s(\widetilde{\mathbf{X}})
+
\mathrm{TaylorKAN}_s
\left(
\mathrm{LN}
\left(
\mathcal{P}_s(\widetilde{\mathbf{X}})
\right)
\right),
\label{eq:skan_token}
\end{equation}
where $\mathbf{U}_s\in\mathbb{R}^{L_p\times d_h\times G}$. We restore the original group-feature order by
\begin{equation}
\overline{\mathbf{U}}_s
=
\mathcal{P}_s^{-1}(\mathbf{U}_s)
\in
\mathbb{R}^{L_p\times G\times d_h}.
\label{eq:skan_restore}
\end{equation}
The channel mixer remains an MLP applied to the feature dimension:
\begin{equation}
\mathbf{V}_s
=
\overline{\mathbf{U}}_s
+
\mathbf{W}_2
\sigma
\left(
\mathbf{W}_1
\mathrm{LN}
(\overline{\mathbf{U}}_s)
+
\mathbf{b}_1
\right)
+
\mathbf{b}_2,
\label{eq:skan_channel}
\end{equation}
where $\mathbf{V}_s\in\mathbb{R}^{L_p\times G\times d_h}$ and $\sigma$ denotes the GELU activation. The node-level spatial output is
\begin{equation}
\mathbf{S}_{\tau,n,c}
=
\mathbf{X}_{p,\tau,n,c}
+
\sum_{g=1}^{G}
\mathbf{G}_{n,g}
\mathbf{V}_{s,\tau,g,c}.
\label{eq:skan_node_restore}
\end{equation}

\textbf{Temporal KAN Blocks.} 
The TKAN block applies TaylorKAN token mixing along the compressed temporal dimension. We define a temporal permutation operator $\mathcal{P}_t:\mathbb{R}^{L_p\times N\times d_h}\rightarrow\mathbb{R}^{N\times d_h\times L_p}$.
The temporal token mixer is
\begin{equation}
\mathbf{U}_t
=
\mathcal{P}_t(\mathbf{S})
+
\mathrm{TaylorKAN}_t
\left(
\mathrm{LN}
\left(
\mathcal{P}_t(\mathbf{S})
\right)
\right),
\label{eq:tkan_token}
\end{equation}
where $\mathbf{U}_t\in\mathbb{R}^{N\times d_h\times L_p}$. The channel mixer is written as
\begin{equation}
\mathbf{V}_t
=
\mathcal{P}_t^{-1}(\mathbf{U}_t)
+
\mathrm{MLP}_t
\left(
\mathrm{LN}
\left(
\mathcal{P}_t^{-1}(\mathbf{U}_t)
\right)
\right),
\label{eq:tkan_channel}
\end{equation}
and the final temporal representation is
\begin{equation}
\mathbf{T}
=
\mathbf{V}_t
\in
\mathbb{R}^{L_p\times N\times d_h}.
\label{eq:tkan_output}
\end{equation}

\subsection{Spatio-Temporal Fusion Blocks}

To complement the TaylorKAN token mixers, we introduce Transformer layers along both spatial and temporal axes. Given a hidden representation $\mathbf{T}\in\mathbb{R}^{L_p\times N\times d_h}$, the attention block projects it into query, key, and value matrices:
\begin{equation}
\mathbf{Q} = \mathbf{T}\mathbf{W}_Q,\quad
\mathbf{K} = \mathbf{T}\mathbf{W}_K,\quad
\mathbf{V} = \mathbf{T}\mathbf{W}_V,
\end{equation}
where $\mathbf{W}_Q,\mathbf{W}_K,\mathbf{W}_V\in\mathbb{R}^{d_h\times d_h}$. The scaled dot-product attention is
\begin{equation}
\mathrm{Attention}(\mathbf{Q},\mathbf{K},\mathbf{V})
=
\mathrm{Softmax}
\left(
\frac{\mathbf{Q}\mathbf{K}^{\top}}
{\sqrt{d_k}}
\right)
\mathbf{V}.
\label{eq:attention}
\end{equation}
Spatial attention is computed independently at each time step along the node dimension, so the softmax operates over the $N$ spatial nodes. Temporal attention is computed independently at each node along the compressed temporal dimension, so the softmax operates over the $L_p$ temporal tokens. Figure~\ref{fig:1} indicates that the attention block contains LayerNorm, FFN, and residual connections in addition to the attention operation.

\subsection{Prediction Head}
After feature extraction and fusion via STKAN blocks, the prediction head aggregates temporal information across compressed time steps for each node and applies a linear projection to generate multi-step forecasts. Let $\mathbf{Z}$ denote the output representation after the fusion block. The prediction head is expressed as:

\begin{equation}
\widehat{\mathbf{Y}}_{t+1:t+H}
=
f_{\mathrm{out}}(\mathbf{Z}),
\label{eq:stid_prediction}
\end{equation}
where $\widehat{\mathbf{Y}}_{t+1:t+H}\in\mathbb{R}^{H\times N\times C_{\mathrm{out}}}$. The output function $f_{\mathrm{out}}$ denotes the reshape and linear projection operations used to map the fused representation to the future horizon.

\section{Experiment}


\subsection{Experimental Setup}
\textbf{Datasets.} We evaluate our model on five traffic forecasting datasets, including PEMS04, PEMS07, PEMS08, PEMS-BAY and METR-LA. Following previous work, we divide the PEMS04, PEMS07 and PEMS08 dataset into training, validation, and test sets in a ratio of 6:2:2. For the remaining datasets, we adopt a split ratio of 7:1:2. Detailed statistics of these datasets are shown in Table \ref{tab:Dataset}.
\begin{table}[t]
\centering
\small
\renewcommand{\arraystretch}{1.1}
\resizebox{\columnwidth}{!}{%
\begin{tabular}{l|ccccc}
\toprule
\textbf{Dataset} & \textbf{PEMS04} & \textbf{PEMS07} & \textbf{PEMS08} & \textbf{PEMS-BAY} & \textbf{METR-LA} \\
\midrule
Sensors          & 307             & 883             & 170             & 325               & 207              \\
Time Steps       & 16992           & 28224           & 17856           & 52116             & 34272            \\
Time Interval    & 5min            & 5min            & 5min            & 5min              & 5min             \\
\bottomrule
\end{tabular}%
}
\caption{Summary of Five Spatio-temporal Benchmarks }
\label{tab:Dataset}
\end{table}

\begin{sloppypar}
\textbf{Baselines.} 
We compare 11 representative baselines with our proposed STKAN. 
(i) \textbf{Non-spatial modeling-based:} STID~\cite{shao2022spatial}, which adopts identity spatio-temporal embeddings and avoids explicit spatial dependency modeling. 
(ii) \textbf{Static spatial-based methods:} STGCN~\cite{yu2017spatio}, GWNet~\cite{wu2019graph}, AGCRN~\cite{bai2020adaptive}, GMAN~\cite{zheng2020gman}, MTGNN~\cite{wu2020connecting} and STDN~\cite{STDN} combine pre-defined or learned static graph structures with temporal modeling modules. 
(iii) \textbf{Dynamic spatial-based methods:} STAEformer~\cite{staeformer} and STWave~\cite{stwave} capture time-varying spatial dependencies through adaptive or attention-based mechanisms. 
(iv) \textbf{Spatio-temporal decomposition-based:} StemGNN~\cite{stemgnn} and STNorm~\cite{deng2021st} decompose spatio-temporal series into separate components for modeling, focusing on disentangling spatial and temporal patterns.
\end{sloppypar}

\textbf{Settings.}  In the experiments, we use the traffic flow of the last 12 time steps to predict the traffic flow of the next 12 time steps, and record the prediction performance of the 3rd, 6th, 12th steps and the average. We set the Adam optimizer with an initial learning rate of 0.002, where the learning rate follows a step-wise decay strategy, and the batch size is set as 64. During the training phase, we employ the early stopping strategy with tolerance 30 for 200 epochs. For performance evaluation, we adopt three widely used metrics to quantify the accuracy of traffic forecasting results: Mean Absolute Error (MAE), Root Mean Squared Error (RMSE), and Mean Absolute Percentage Error (MAPE). All baseline methods are implemented within the BasicTS \cite{BasicTS} framework to ensure consistent training pipelines and fair performance comparison.

\begin{table*}[!t]
\renewcommand\arraystretch{1}
    \centering
    \label{tab:main}
    \resizebox{\textwidth}{!}{
    \begin{tabular}{lc|cccc|cccc|cccc|cccc|cccc}
    \toprule
    \midrule
    \multicolumn{2}{c|}{\textbf{Dataset}} & \multicolumn{4}{c|}{\textbf{PEMS04}} & \multicolumn{4}{c|}{\textbf{PEMS07}} & \multicolumn{4}{c|}{\textbf{PEMS08}}& \multicolumn{4}{c|}{\textbf{PEMS-BAY}} & \multicolumn{4}{c}{\textbf{METR-LA}}\tabularnewline
    \midrule
    \midrule
    \textbf{Method} & \textbf{Metric} & \textbf{@3} & \textbf{@6} & \textbf{@12} & \textbf{Avg.} & \textbf{@3} & \textbf{@6} & \textbf{@12} & \textbf{Avg.} & \textbf{@3} & \textbf{@6} & \textbf{@12} & \textbf{Avg.} & \textbf{@3} & \textbf{@6} & \textbf{@12} & \textbf{Avg.} & \textbf{@3} & \textbf{@6} & \textbf{@12} & \textbf{Avg.}\tabularnewline
    \midrule
\multirow{3}{*}{GWNet (2019)} & MAE & 17.89 & 18.80 & 20.35 & 18.81 & 18.71 & 20.14 & 22.35 & 20.10 & 13.67 & 14.59 & 15.99 & 14.58 & 1.31 & 1.65 & 1.99 & 1.59 & 2.69 & 3.08 & 3.52 & 3.04 \\
& RMSE & 28.81 & 30.40 & 32.66 & 30.38 & 30.70 & 33.20 & 36.58 & 33.11 & 21.65 & 23.54 & 25.82 & 23.46 &  \underline{2.76} & 3.74 & 4.54 & 3.66 & 5.17 & 6.20 & 7.28 & 6.15 \\
& MAPE & 12.23\% & 12.99\% & 14.24\% & 12.97\% & 8.04\% & 8.50\% & 9.73\% & 8.59\% & 9.20\% & 9.69\% & 10.41\% & 9.69\% & 2.77\% & 3.80\% & 4.84\% & 3.66\% & 6.93\% & 8.33\% & 9.84\% & 8.15\% \\
    \midrule
\multirow{3}{*}{STGCN (2018)} & MAE & 19.09 & 19.98 & 21.74 & 20.03 & 20.67 & 22.23 & 25.04 & 22.28 & 15.97 & 16.86 & 18.64 & 16.96 & 1.41 & 1.75 & 2.08 & 1.70 & 2.76 & 3.15 & 3.63 & 3.12 \\
& RMSE & 30.1 & 31.57 & 34.07 & 31.63 & 32.76 & 35.71 & 40.4 & 35.83 & 24.56 & 26.29 & 29.00 & 26.38 & 2.93 & 3.89 & 4.69 & 3.81 & 5.30 & 6.32 & 7.47 & 6.29 \\
& MAPE & 12.95\% & 13.44\% & 14.73\% & 13.73\% & 8.97\% & 9.53\% & 10.71\% & 9.59\% & 10.91\% & 11.56\% & 12.58\% & 11.50\% & 3.06\% & 3.98\% & 4.85\% & 3.84\% & 7.11\% & 8.61\% & 10.40\% & 8.49\% \\
\midrule
\multirow{3}{*}{AGCRN (2020)} & MAE & 18.55 & 19.50 & 20.77 & 19.45 & 19.29 & 20.82 & 22.81 & 20.74 & 14.78 & 15.96 & 17.63 & 15.91 & 1.35 & 1.67 & 1.96 & 1.61 & 2.87 & 3.24 & 3.63 & 3.19 \\
& RMSE & 29.86 & 31.60 & 33.51 & 31.46 & 31.64 & 34.64 & 38.05 & 34.50 & 22.98 & 25.01 & 27.79 & 25.03 & 2.85 & 3.80 & 4.54 & 3.69 & 5.61 & 6.66 & 7.58 & 6.52 \\
& MAPE & 12.88\% & 13.44\% & 14.18\% & 13.40\% & 8.15\% & 8.70\% & 9.70\% & 8.83\% & 9.53\% & 11.72\% & 12.17\% & 10.86\% & 2.93\% & 3.84\% & 4.68\% & 3.69\% & 7.74\% & 9.03\% & 10.30\% & 8.85\% \\
\midrule
\multirow{3}{*}{StemGNN (2020)} & MAE & 19.14 & 20.82 & 24.05 & 21.00 & 20.78 & 23.25 & 27.91 & 23.41 & 14.63 & 16.05 & 18.76 & 16.20 & 1.39 & 1.78 & 2.20 & 1.73 & 2.97 & 3.50 & 4.24 & 3.49 \\
& RMSE & 30.38 & 32.78 & 37.09 & 33.06 & 32.78 & 36.56 & 43.05 & 36.89 & 22.98 & 25.42 & 29.45 & 25.62 & 2.92 & 3.95 & 4.94 & 3.90 & 5.82 & 7.04 & 8.59 & 7.06 \\
& MAPE & 13.68\% & 14.82\% & 17.44\% & 15.05\% & 9.29\% & 10.21\% & 12.45\% & 10.39\% & 9.28\% & 10.50\% & 12.26\% & 10.55\% & 2.94\% & 4.09\% & 5.32\% & 3.97\% & 7.97\% & 10.06\% & 13.01\% & 10.04\% \\
\midrule
\multirow{3}{*}{GMAN (2020)} & MAE & 18.23 & 18.78 & 20.12 & 18.81 & 19.31 & 20.41 & 22.20 & 20.48 & 13.76 & 14.59 & 15.83 & 14.81 & 1.35 & 1.66 & 1.93 & 1.58 & 2.81 & 3.15 & 3.49 & 3.07 \\
& RMSE & 29.38 & 30.91 & 31.25 & 30.99 & 31.25 & 33.32 & 36.51 & 33.40 & 22.78 & 24.15 & 26.49 & 24.23 & 2.92 & 3.84 & 4.51 & 3.69 & 5.56 & 6.50 & 7.36 & 6.43 \\
& MAPE & 12.71\% & 13.27\% & 13.41\% & 13.22\% & 8.22\% & 8.71\% & 9.44\% & 8.65\% & 9.40\% & 9.53\% & 10.56\% & 9.71\% & 2.88\% & 3.75\% & 4.54\% & 3.69\% & 7.42\% & 8.75\% & 10.11\% & 8.65\% \\
\midrule
\multirow{3}{*}{MTGNN (2020)} & MAE & 18.29 & 19.12 & 20.57 & 19.12 & 19.52 & 21.11 & 23.87 & 21.16 & 14.23 & 15.30 & 16.97 & 15.31 & 1.32 & 1.65 & 1.95 & 1.59 & 2.70 & 3.07 & 3.53 & 3.04 \\
& RMSE & 29.82 & 31.34 & 33.57 & 31.28 & 31.37 & 34.19 & 38.46 & 34.26 & 22.38 & 24.33 & 26.78 & 24.25 & 2.78 & 3.73 & 4.50 & \underline{3.65} & 5.21 & 6.17 & 7.24 & 6.14 \\
& MAPE & 12.62\% & 13.09\% & 14.31\% & 13.14\% & 8.77\% & 9.10\% & 10.34\% & 9.27\% & 9.42\% & 10.57\% & 12.17\% & 10.60\% & 2.75\% & \underline{3.68\%} & 4.55\% & \underline{3.53\%} & 6.85\% & 8.17\% & 9.81\% & 8.08\% \\
\midrule
\multirow{3}{*}{STNorm (2021)} & MAE & 18.30 & 19.12 & 20.27 & 19.05 & 19.21 & 20.57 & 22.66 & 20.51 & 14.48 & 15.45 & 17.03 & 15.45 & 1.33 & 1.66 & 1.97 & \underline{1.58} & 2.80 & 3.18 & 3.56 & 3.12 \\
& RMSE & 29.82 & 31.52 & 33.22 & 31.28 & 31.65 & 34.66 & 38.30 & 34.48 & 23.05 & 25.38 & 27.93 & 25.22 & 2.85 & 3.81 & 4.56 & 3.67 & 5.49 & 6.52 & 7.47 & 6.41 \\
& MAPE & 12.32\% & 12.83\% & 13.69\% & 12.81\% & 8.29\% & 8.69\% & 9.61\% & 8.70\% & 9.27\% & 9.79\% & 10.90\% & 9.88\% & 2.85\% & 3.77\% & 4.63\% & 3.59\% & 7.44\% & 8.89\% & 10.26\% & 8.65\% \\
\midrule
\multirow{3}{*}{STID (2022)} & MAE & 17.62 & 18.40 & 19.72 & 18.41 & 18.40 & 19.66 & 21.54 & 19.62 & \underline{13.29} & \underline{14.22} & \underline{15.55} & \underline{14.20} & {1.31} & \underline{1.64} & \underline{1.91} & 1.56 & 2.79 & 3.17 & 3.54 & 3.11 \\
& RMSE & \textbf{28.61} & \textbf{29.95} & 31.93 & \underline{29.93} & \underline{30.45} & \textbf{32.82} & \textbf{36.04} & \underline{32.75} & \underline{21.53} & \underline{23.40} & \underline{25.72} & \underline{23.34} & \underline{2.77} & \underline{3.73} & \underline{4.40} & 3.60 & 5.52 & 6.57 & 7.53 & 6.47 \\
& MAPE & 11.95\% & 12.42\% & 13.50\% & 12.51\% & 7.77\% & 8.28\% & 9.22\% & 8.31\% & \underline{8.65\%} & \underline{9.29\%}& \underline{10.32\%} & \underline{9.31\%} & 2.77\% & 3.73\% & \underline{4.52\%} & 3.55\% & 7.66\% & 9.27\% & 10.77\% & 9.01\% \\
\midrule
\multirow{3}{*}{STAEformer (2023)} & MAE & \underline{17.48} & 18.24 & \underline{19.30} & \underline{18.19} & \underline{18.00} & \underline{19.40} & \underline{21.42} & \underline{19.33} & \underline{12.71} & \underline{13.55} & \underline{14.84} & \underline{13.55} & \underline{1.30} & \textbf{1.61} & \textbf{1.87} & \textbf{1.54} & \textbf{2.65} & \textbf{2.96} & \textbf{3.33} & \textbf{2.93} \\
& RMSE & 28.89 & 30.31 & 31.99 & 30.18 & 30.42 & 33.30 & 37.02 & 33.21 & 21.63 & 23.48 & 25.80 & 23.44 & \underline{2.77} & \textbf{3.68} & \underline{4.34} & \underline{3.57} & \textbf{5.11} & \textbf{6.01} & \textbf{7.02} & \textbf{5.98} \\
& MAPE & 11.78\% & \underline{12.21\%} & \underline{13.00\%} & 12.25\% & \underline{7.61\%} & 8.19\% & \underline{9.03\%} & \underline{8.14\%} & \underline{8.33\%} & \underline{8.92\%} & \underline{9.85\%} & \underline{8.90\%} & \underline{2.74\%} & \underline{3.63\%} & \underline{4.41\%} & \underline{3.46\%} & \textbf{6.90\%} & \textbf{8.20\%} & \underline{9.77\%} & \textbf{8.10\%} \\
\midrule
\multirow{3}{*}{STWave (2023)} & MAE & 17.57 & \underline{18.17} & 19.42 & 18.25 & 18.57 & 19.91 & 21.75 & 19.93 & 12.78 & 13.76 & 14.86 & 13.69 & 1.32 &  \underline{1.63} &  \underline{1.89} &  \underline{1.56} & 2.83 & 3.22 & 3.58 & 3.15 \\
& RMSE & 28.88 & \textbf{29.95} & \underline{31.78} & 29.99 & 31.59 & 34.36 & 37.35 & 34.09 & 21.59 & 23.79 & 25.77 & 23.57 & 2.80 & \underline{3.71} & 4.35 & 3.59 & 5.63 & 6.71 & 7.60 & 6.56 \\
& MAPE & \textbf{11.65\%} & \textbf{12.02\%} & 13.13\% & \textbf{12.16\%} & 7.63\% & \underline{8.17\%} & 9.07\% & 8.20\% & 8.63\% & 9.16\% & 10.03\% & 9.10\% & 2.76\% & 3.66\% & 4.44\% & 3.50\% & 7.72\% & 9.49\% & 11.03\% & 9.20\% \\
\midrule
\multirow{3}{*}{STDN (2025)} & MAE & 18.15 & 18.89 & 20.14 & 18.92
& 19.92 & 21.16 & 23.51 & 21.29 
& 13.85 & 14.43 & 15.71 & 14.53 
& 1.38 & 1.66 & 1.93 & 1.61 
& 2.79 & 3.15 & 3.53 & 3.10 \\
& RMSE & 33.14 & 34.64 & 35.85 & 34.33
& 33.56 & 35.88 & 39.61 & 36.02 
& 22.31 & 23.90 & 26.21 & 23.96 
& 2.95 & 3.83 & 4.47 & 3.66 
& 5.59 & 6.61 & 7.56 & 6.51 \\
& MAPE & 19.34\% & 19.24\% & 19.80\% & 19.22\% 
& 12.73\% & 12.12\% & 14.78\% & 12.85\% 
& 12.45\% & 11.64\% & 10.92\% & 11.43\% 
& 3.03\% & 3.81\% & 4.47\% & 3.66\% 
& 7.61\% & 9.08\% & 10.73\% & 8.93\% \\
\midrule
\multirow{3}{*}{\textbf{STKAN(Ours)}} 
& MAE & \textbf{17.40}  &  \textbf{18.13} & \textbf{19.15}  & \textbf{18.09}  &
\textbf{17.94} &  \textbf{19.27} &  \textbf{20.97} &  \textbf{19.16} & 
\textbf{12.62} & \textbf{13.48} & \textbf{14.80} & \textbf{13.45} &
\textbf{1.29}  & \textbf{1.61}  & \textbf{1.87}  & \textbf{1.54}  & 
\underline{2.69}  & \underline{2.99}  & \underline{3.39}  & \underline{2.97}  \\
& RMSE & \underline{28.63}  &  \underline{30.00} & \textbf{31.59}  & \textbf{29.87}  &  
\textbf{30.13} &  \underline{32.96} &  \underline{36.12} & \textbf{32.74}  & 
\textbf{21.25} & \textbf{23.24} & \textbf{25.56} &  \textbf{23.17} &  
 \textbf{2.73} & \textbf{3.68}  & \textbf{4.33}  &  \textbf{3.56} &  
\underline{5.13} & \underline{6.08}  &  \underline{7.14} & \underline{6.06}  \\
& MAPE & \underline{11.78\%} &  12.24\% &  \textbf{12.95\%}&  \underline{12.23\%} &  
\textbf{7.60\%} &  \textbf{8.05\%} &  \textbf{8.93\%} &  \textbf{8.07\%} & 
\textbf{8.28\%} & \textbf{8.90\%} & \textbf{9.84\%} &  \textbf{8.88\%} &
\textbf{2.70\%} &  \textbf{3.62\%} &  \textbf{4.35\%} &  \textbf{3.44\%} &
\underline{6.98\%} &  \underline{8.25\%} &  \textbf{9.71\%} &  \underline{8.11\%} \\
\midrule
\bottomrule
\end{tabular}}
\caption{Forecasting performance on the five benchmark datasets. We bold the best results and underline the second-best results.}
\label{tab:normal}
\end{table*}

\subsection{Performance Comparisons}
Table~\ref{tab:normal} reports the forecasting results on the five benchmark datasets. STKAN achieves the best average MAE and RMSE on PEMS04, although its average MAPE is slightly higher than that of STWave. On PEMS07 and PEMS08, STKAN obtains the best average results across the three reported metrics. On PEMS-BAY, STKAN ties with STAEformer in average MAE and obtains the lowest average RMSE and MAPE. On METR-LA, STKAN remains competitive but does not outperform the strongest baseline in average MAE or RMSE. Overall, the results indicate that STKAN is particularly effective on the evaluated traffic-flow datasets, whereas its advantage is less evident on the METR-LA speed dataset.

The gains over the best-performing baselines are generally modest. Since the current evaluation reports single-run results, small numerical differences should be interpreted cautiously. These results demonstrate the competitiveness of the overall STKAN architecture. The comparison with the MLP variant further suggests that the adopted TaylorKAN token mixers are useful within the evaluated configuration, although the current experiments do not isolate function-approximator capacity as the sole source of the observed gains.

\subsection{Ablation Study}

\textbf{Effectiveness of KAN Modules.} 
To evaluate the role of KAN components within STKAN, we design three model variants: \begin{itemize} \item \textbf{MLPs}: replacing the TaylorKAN token mixers with MLP layers in the evaluated implementation. \item \textbf{w/o SKAN}: removing the spatial block while retaining TKAN to test the importance of inter-node mixing. \item \textbf{w/o TKAN}: removing the temporal block while retaining the spatial modeling component. \end{itemize}

\begin{figure}[t]
  \centering
  \includegraphics[width=\linewidth]{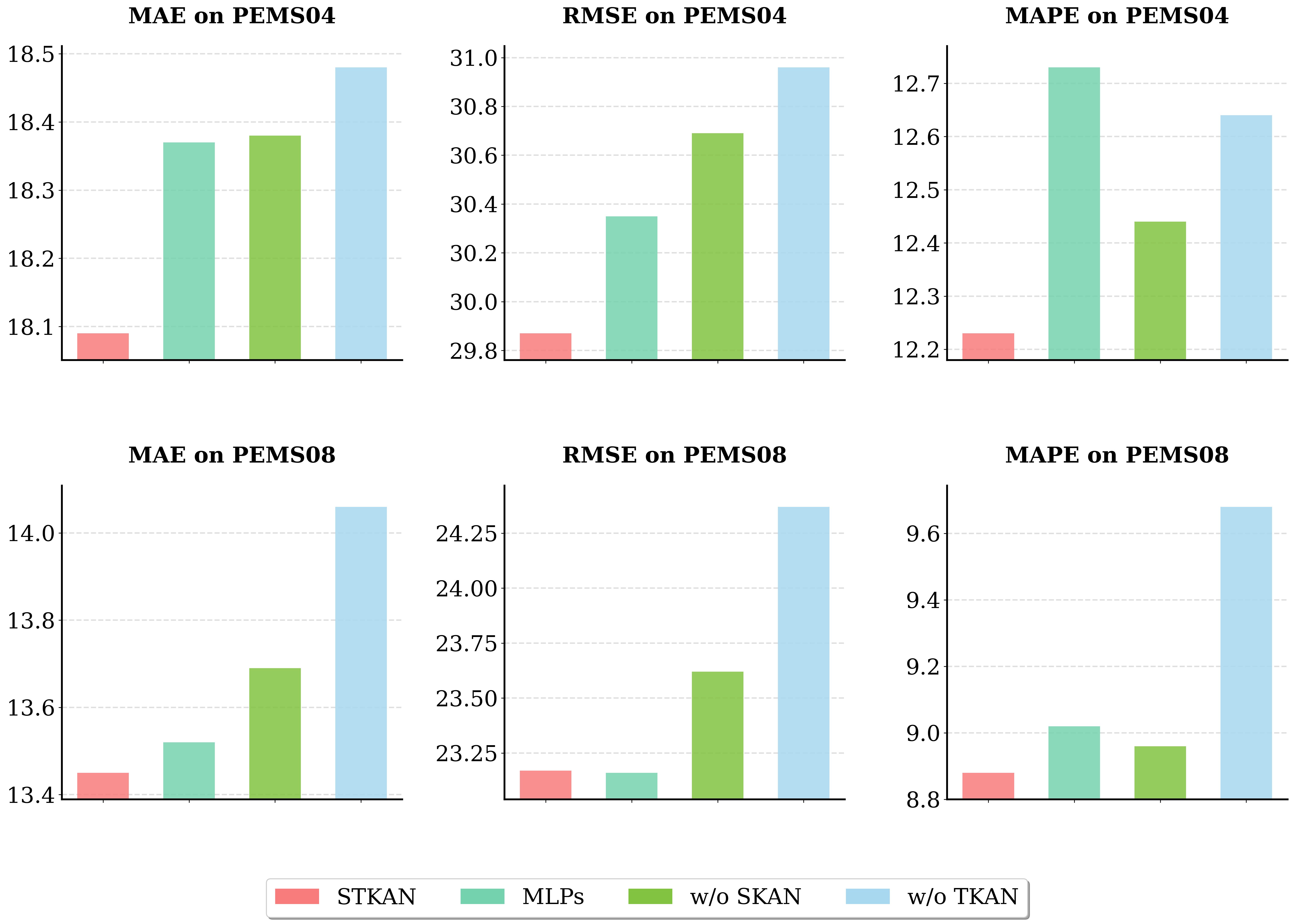}
  \caption{KAN modules ablation on PEMS04 and PEMS08.}
  \label{fig:3}
\end{figure}

As shown in Figure~\ref{fig:3}, replacing the TaylorKAN token mixers with the evaluated MLP implementation increases the forecasting errors on PEMS04 and PEMS08. This result suggests that the Taylor-polynomial mappings contribute positively within the current architecture and hyperparameter setting. Removing either SKAN or TKAN also degrades at least part of the reported performance, indicating that spatial-group mixing and temporal mixing provide complementary information. These observations are specific to the evaluated model configurations and should not be interpreted as establishing the general advantage of KANs over all parameter-matched MLP alternatives.

\textbf{Effectiveness of Attention Mechanisms.} To evaluate the role of attention in STKAN, we conduct ablation studies on PEMS04 and PEMS08 with the following variants:  
\begin{itemize} \item \textbf{w/o S-Attention}: disabling spatial attention while preserving temporal modeling. \item \textbf{w/o T-Attention}: removing temporal attention while keeping spatial modeling. \item \textbf{w/o ST-Attention}: removing both attention modules, leaving only KAN-based token and channel mixing. \end{itemize}

Table~\ref{tab:stkan_ablation2} shows that the complete model achieves the best result for every reported metric on PEMS04 and PEMS08. However, the relative influence of spatial and temporal attention varies across metrics. For example, removing temporal attention produces the largest MAE degradation on PEMS08, whereas removing spatial attention produces the largest RMSE degradation. Removing both branches is therefore not uniformly the worst variant for every metric. The results indicate that the two attention branches contribute differently across datasets and evaluation criteria.

The ablation results suggest that attention provides an additional refinement over the KAN-based mixer blocks, but the current experiments do not establish a strict ranking between the contributions of attention and KAN components.

\begin{table}[t]
  \centering
  \small
  \setlength{\tabcolsep}{3pt}
    \scalebox{0.95}{
    \begin{tabular}{c|ccc|ccc}
    \toprule
    Dataset & \multicolumn{3}{c|}{PEMS04} & \multicolumn{3}{c}{PEMS08} \\
    \midrule
    Metric  & MAE & RMSE & MAPE & MAE & RMSE & MAPE \\
    \midrule
    w/o S-Attention & 18.24 & 30.13 & 12.43\% & 13.70 & 23.68 & 9.06\% \\
    w/o T-Attention & 18.28 & 30.27 & 12.40\% & 13.84 & 23.39 & 9.20\% \\
    w/o ST-Attention& 18.31 & 30.05 & 12.52\% & 13.57 & 23.30 & 8.99\% \\
    \textbf{STKAN}  & \textbf{18.09} & \textbf{29.87} & \textbf{12.23\%} &
                      \textbf{13.45} & \textbf{23.17} & \textbf{8.88\%} \\
    \bottomrule
    \end{tabular}
  }
    \caption{Ablation study of the attention block.}
  \label{tab:stkan_ablation2}

\end{table}

\subsection{Hyper-parameter Study}

Table~\ref{tab:g} reports the sensitivity of STKAN to the number of spatial groups $G$. The results exhibit a non-monotonic pattern: using either fewer or more groups than the selected value leads to slightly higher forecasting errors. Among the tested settings, $G=16$ performs best on PEMS04, while $G=20$ performs best on PEMS07. This observation suggests that the group number should be selected according to the dataset rather than increased monotonically with the number of nodes.

\begin{table}[htbp]
\centering
\renewcommand{\arraystretch}{1.2}
\setlength{\tabcolsep}{4pt} 

\small 
\begin{tabular}{c|ccc|c|ccc}
\toprule
\multicolumn{4}{c|}{PEMS04} & \multicolumn{4}{c}{PEMS07} \\ \hline
$G$ & MAE & RMSE & MAPE & $G$ & MAE & RMSE & MAPE \\ \hline
8  & 18.28 & 30.76 & 12.55\% & 12 & 19.41 & 33.18 & 8.19\% \\
12 & 18.20 & 30.16 & 12.40\% & 16 & 19.18 & 32.77 & 8.10\% \\
16 & \textbf{18.09} & \textbf{29.87} & \textbf{12.23\%} & 20 & \textbf{19.16} & \textbf{32.74} & \textbf{8.08\%} \\
20 & 18.20 & 30.00 & 12.44\% & 24 & 19.26 & 32.86 & 8.17\% \\
24 & 18.25 & 30.25 & 12.61\% & 28 & 19.24 & 32.93 & 8.13\% \\ 
\bottomrule
\end{tabular}

\caption{Hyper-parameter study on varying $G$ values.}
\label{tab:g}
\end{table}

\subsection{Case Study}

\begin{figure}[t]
  \centering
  \includegraphics[width=0.9\linewidth]{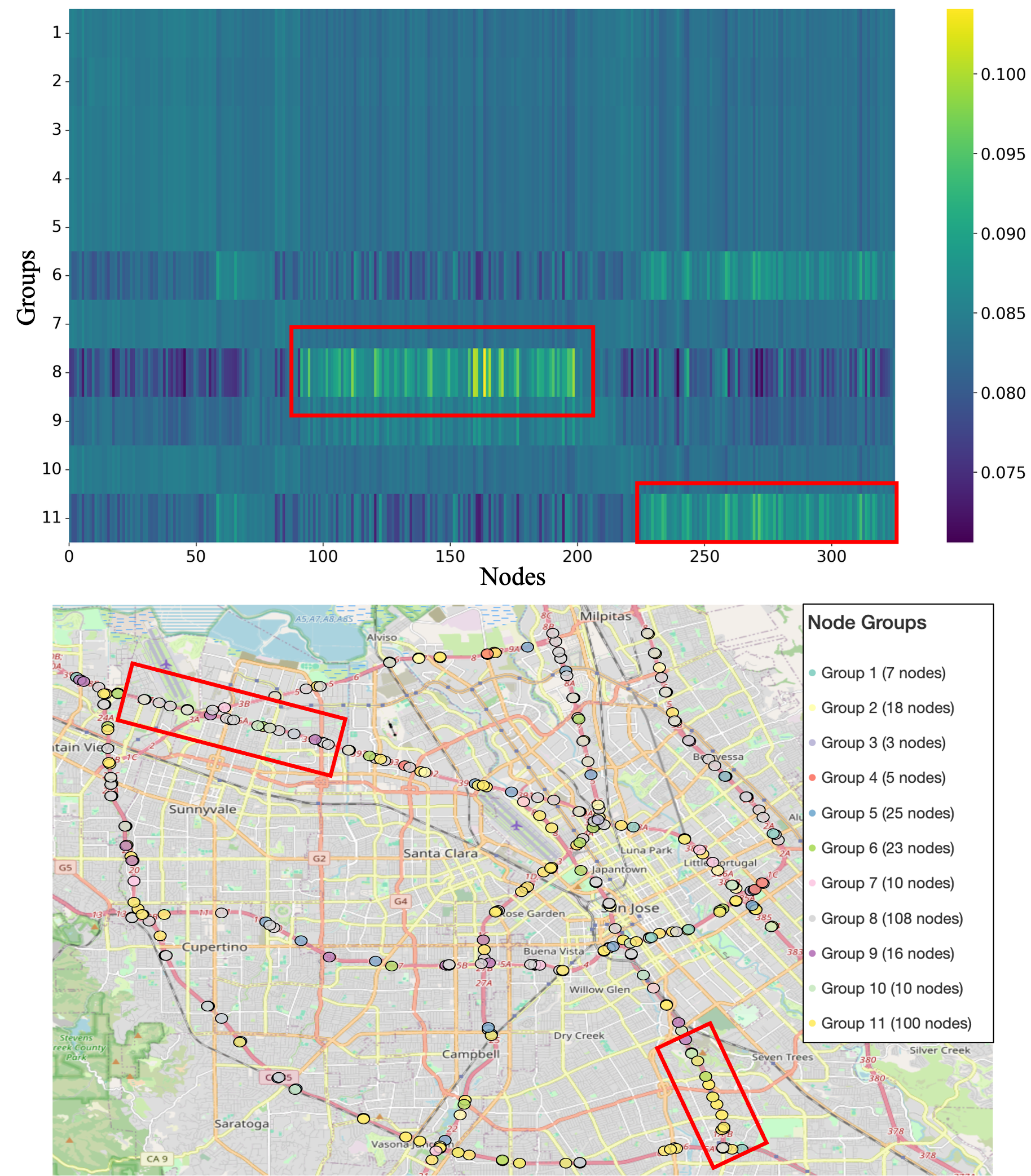}
  \caption{Adaptive grouping matrices visualization on PEMS-BAY.}
  \label{fig:group_assignment}
\end{figure}

Figure~\ref{fig:group_assignment} provides a qualitative visualization of the learned soft node-to-group assignment matrix on PEMS-BAY. Several groups exhibit relatively concentrated assignment patterns, and some corresponding hard assignments appear to cover spatially contiguous road segments. This observation suggests that the learned grouping mechanism may capture corridor-level regularities in the sensor network. However, the visualization should be interpreted as qualitative evidence rather than a formal validation of geographical consistency or model interpretability.

\section{Conclusion}
In this work, we introduced STKAN, a spatio-temporal forecasting architecture that incorporates Taylor-polynomial KAN mappings into spatial and temporal token-mixing modules. The model combines learnable soft node grouping, group-wise spatial mixing, temporal mixing, and spatial--temporal attention to model traffic dynamics. Experiments on five benchmark datasets show that STKAN achieves competitive forecasting performance, with particularly strong results on the evaluated traffic-flow datasets. The comparison with the tested MLP variant suggests that TaylorKAN token mixers can provide a useful alternative nonlinear parameterization within the proposed architecture. These findings indicate that function-approximator design is a relevant component of spatio-temporal modeling, alongside spatial and temporal architectural design. Future work may examine parameter-matched comparisons, computational efficiency, statistical variability, and broader spatio-temporal applications.

\bibliographystyle{ACM-Reference-Format}
\bibliography{ijcai26}

\end{document}